%% file: MAIN.tex
\documentclass{article}
\usepackage{ijcai25}

\usepackage{times}
\usepackage{soul}
\usepackage{url}
\usepackage[hidelinks]{hyperref}
\usepackage[utf8]{inputenc}
\usepackage[small]{caption}
\usepackage{graphicx}
\usepackage{amsmath}
\usepackage{amssymb}
\usepackage{amsthm}
\usepackage{booktabs}
\usepackage{algorithm}
\usepackage{algorithmic}
\usepackage[switch]{lineno}
\usepackage{color}
\usepackage[utf8]{inputenc}
\usepackage{svg}
\usepackage{multirow}
\usepackage{PTSansNarrow}

\newcommand{\name}{CHECK }
\newcommand{\nameNoSpace}{CHECK}

\title{Knowledge Editing for Multi-Hop Question Answering Using Semantic Analysis}

\author{
Dominic Simon
\and
Rickard Ewetz
\affiliations
University of Florida\\
\emails
\{dominic.simon, rewetz\}@ufl.edu,
}
\begin{document}

\maketitle

\begin{abstract} 
Large Language Models (LLMs) require lightweight avenues of updating stored information that has fallen out of date. Knowledge Editing (KE) approaches have been successful in updating model knowledge for simple factual queries but struggle with handling tasks that require compositional reasoning such as multi-hop question answering (MQA). We observe that existing  knowledge editors leverage decompositional techniques that result in illogical reasoning processes. In this paper, we propose a knowledge editor for MQA based on semantic analysis called \nameNoSpace. Our framework is based on insights from an analogy between compilers and reasoning using LLMs. Similar to how source code is first compiled before being executed, we propose to semantically analyze reasoning chains before executing the chains to answer questions. Reasoning chains with semantic errors are revised to ensure consistency through logic optimization and re-prompting the LLM model at a higher temperature. We evaluate the effectiveness of \name against five state-of-the-art frameworks on four datasets and achieve an average 22.8\% improved MQA accuracy.
\end{abstract}

\section{Introduction}
Large Language Models (LLM) are trained on extensive amounts of data, enabling them to grasp the statistical patterns of natural language and broad factual knowledge~\cite{gpt3}. The factual knowledge is utilized when  LLMs are integrated into applications such as chatbots, translators, and question-answering systems~\cite{translation}. It is unavoidable that the factual knowledge stored within LLMs becomes outdated over time. Retraining the LLMs from scratch to learn new factual data is both economically expensive~\cite{llm_cost} and introduces an undue burden on the environment~\cite{co2}. The concept of knowledge editing (KE) has emerged as a promising solution to bypass the need for retraining LLMs from scratch. Knowledge editing approaches commonly fall into two categories:  parameter-based approaches that inject edits directly into model parameters~\cite{rome,memit,unbias,common_sense,believe} and memory-based methods that introduce additional parameters for edit injection~\cite{serac,gpm,melo,wise}. Both of these solution strategies have demonstrated success for basic questioning answering problems~\cite{memit,serac}. However, the problem becomes immensely more challenging when the questions involve compositional reasoning, such as multi-hop question answering (MQA), where the intermediate  knowledge between a hop could have been edited. For example, \emph{What is the country of citizenship of the author of Harry Potter}, where an edit could have changed the \emph{country of citizenship} of \emph{JK Rowling} from \emph{United Kingdom} to \emph{United States}. This emerging challenge has recently spurred further investigations into knowledge editing for MQA problems~\cite{gmello,rae}.  

State-of-the-art knowledge editors for MQA  rely on decomposing the multi-hop problems into multiple single-hop parts~\cite{mquake,pokemqa,deepedit}. The decomposition allows the knowledge editors to compare the intermediate facts with edits stored in a memory bank. The decomposition is performed using an LLM through long in-context examples. However, this approach is prone to leveraging illogical reasoning processes and accidentally utilizing non-relevant edited facts. An intuitive approach to improving existing solutions would be to leverage explicit question decomposition. Nevertheless, explicitly decomposing multi-hop questions into single-hop questions is not straightforward because it may introduce errors from the loss of context, nuances, and hallucinations.

In this paper, we propose a framework for knowledge editing based on semantic analysis called \nameNoSpace. The framework is based on insights from an analogy between compilers and reasoning using LLMs. Source code is first required to pass semantic analysis tests such as type checking before being compiled into a binary that can be executed. Inspired by this approach, we propose to semantically \emph{type check} the reasoning chains generated by LLMs for solving MQA problems. The main contributions of this paper are summarized, as follows: 
\begin{itemize}
    \item We propose the concept of semantically  analyzing the reasoning process of knowledge editors. Each hop in a multi-hop question is assigned a type in the form of person, place, or thing. Next, each of the input and output types within each hop of a reasoning chain are checked for consistency.  

    \item Semantic inconsistencies are resolved by formulating optimization problems to repair the reasoning chains by rearranging the extracted relationships or re-prompting the LLM for a new reasoning chain at a higher temperature.

    \item Experimental evaluation on the MQuAKE dataset demonstrates that \name achieves an average 22.8\% greater accuracy than other similar approaches across three open-source LLMs.
\end{itemize}

The remainder of the paper is broken into the following sections: preliminary knowledge is discussed in Section~\ref{sec:prelim}, the motivation behind using type checking is given in Section~\ref{sec:motivation}, the methodology of the \name framework is explained in Section~\ref{sec:methodology}, experimental results across $4$ datasets and $3$ LLMs are provided in Section~\ref{sec:results}, and the conclusion is in Section~\ref{sec:conclusion}.

\section{Preliminaries}
\label{sec:prelim}
The problem formulation of knowledge editing is provided in Section~\ref{sec:prob_form}. Related works are discussed in Section~\ref{sec:related_works}.

\subsection{Problem Formulation}
\label{sec:prob_form}
This paper addresses the problem of Multi-hop Question Answering (MQA) under Knowledge Editing. A single factual association can be viewed as a subject $s$, relation $r$, object $o$ triple $t=(s, r, o)$, where \textit{Akira Toriyama was born in Japan} can be converted to $(Akira\text{ }Toriyama,\text{ }born\text{ }in,\text{ }Japan)$. Editing a factual relation is updating $o$ to become a new entity $o'$ so that the edited triple becomes $t'=(s, r, o')$. A factual association can be expressed in the form of a question $q=(s, r) \rightarrow o$, where $o$ is unknown until the question is answered. Multi-hop questions $\mathcal{Q}$ contain a chain of relations $\mathcal{Q}=(r_0, r_1, ..., r_{n})$ that can be viewed as a set of subquestions $((s, r_0) \rightarrow o_1, (o_1, r_1) \rightarrow o_2, ..., (o_{n-1}, r_{n-1}) \rightarrow o_n)$ that must be iteratively solved to uncover the obscured entities until the final answer is found. For example, the multi-hop question \textit{Where is the birthplace of the creator of Dragonball?} contains the relations $(creator,\text{ }birthplace)$, which translate into the subquestions $((Dragonball,\text{ }creator)\rightarrow Toriyama,\text{ } (Toriyama,\text{ }birthplace)\rightarrow Japan)$. Answering any of the subquestions $q$ of a multi-hop question $\mathcal{Q}$ with edited information $t'$ will cause the subsequent subquestion answers $o_{n}'$ to deviate from the original answer path, such that $\mathcal{Q}=((s, r_0) \rightarrow o_{1}', (o_{1}', r_1) \rightarrow o_{2}'))$. Generating a correct subquestion path to traverse and determining whether a subquestion requires an edited answer are the two main challenges of MQA under KE.

\subsection{Related Works}
\label{sec:related_works}

In this section, we review studies on knowledge editing for LLMs. Early investigation on KE using parameter-based approaches include~\cite{rome,serac,memit}. However, solution strategies that directly modify model parameters face issues such as catastrophic forgetting~\cite{forget}, one-way associations~\cite{rome,memit}, and long training times~\cite{melo}. Other investigations have focused on augmenting the LLM with external knowledge graphs~\cite{tkemqa,rae,gmello}. However, such solutions are constrained to applications where such graphs are available~\cite{kg_bad}. Embedding-based editors store edits in an embedding space for compact retrieval~\cite{mquake,pokemqa}. MeLLo~\cite{mquake} uses the dense retrieval model Contriever~\cite{contriever} to store factual edit sentences in an embedding space. Next, an in-context learning prompt is used to break the initial question into subquestions and check the subquestion answer against the most similar embedded answer for factual conflicts. PokeMQA~\cite{pokemqa} uses a similar prompting scheme, but it removes the burden of determining conflict from the LLM and trains a two-level conflict disambiguation network to determine whether the subquestion answer and retrieved embedding contain conflicting information. 

The knowledge editing steps included by state-of-the-art frameworks are shown in Table~\ref{tab:program_synthesis}. Neither the above knowledge editors nor similar ones have any way to ensure that the generated subquestions are being answered in an order reflecting the original multi-hop question. This results in misordered chains of subquestions, leading to a question-answering flow that will never arrive at the correct answer. The proposed \name framework resolves this issue by type checking the subquestion reasoning process using semantic analysis.

\input{Tables/program_synthesis}

\begin{figure*}[t]
    \centering
    % \includesvg[width=\textwidth]{Figures/flow.svg}
    \includegraphics[width=\textwidth]{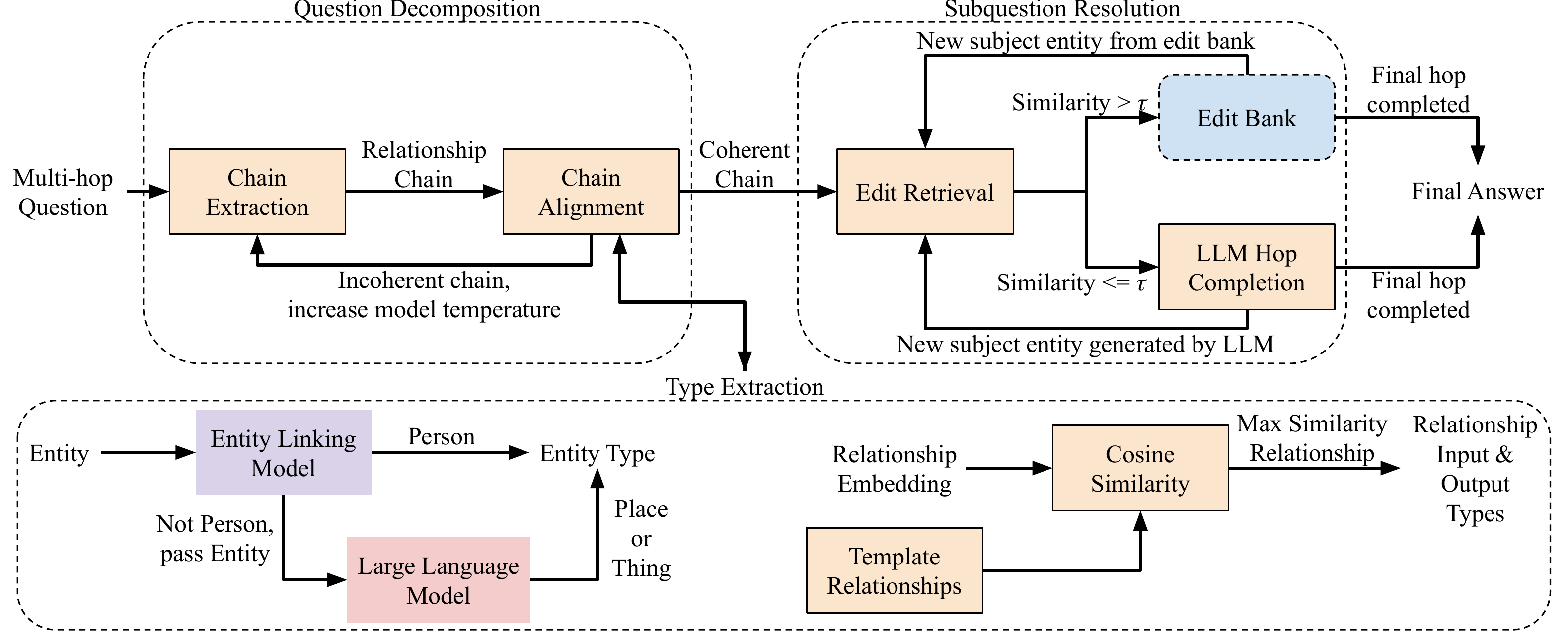}
    \caption{The flow of the proposed \name knowledge editor. \name processes the initial multi-hop question by decomposing it into a chain of relationships. Misaligned chains are realigned through type checking and model temperature increases. Next, each entity and relationship pair is checked to see if an edit is necessary. If an edit is required, the object corresponding to the edit triple is provided as the next entity. Otherwise, the LLM provides the next entity.}
    \label{fig:overview}
\end{figure*}

\section{Semantic Analysis}
\label{sec:motivation}

Our proposed knowledge editing solution is based on insights from an analogy between compilers and reasoning using LLMs. Source code is converted into a binary executable through a compilation process consisting of preprocessing, semantic parsing, assembly conversion, and linking. Next, the binary can be executed to compute an output. Semantic parsing involves type checking to ensure each function call has arguments that match the function definitions. For example, checking that a function expecting an argument of type \texttt{double} is not passed an argument of type \texttt{char}. The compilation process eliminates syntactic and semantic errors, which reduces debugging of the executable to value errors. We propose to adapt this method of semantic analysis to knowledge editing for MQA by type checking the reasoning chains and knowledge edits. The type checking will ensure that the reasoning processes are logical and will assist in eliminating hallucinations from  LLM responses. 
   
In this paper, we propose to categorize all entities as \texttt{persons}, \texttt{places}, or \texttt{things}. Optionally, more fine-grained type categories can be used. 
Consequently, relationships can be viewed as functions that expect inputs and outputs of \texttt{persons}, \texttt{places}, or \texttt{things}. In the sentence \textit{Akira Toriyama was born in Japan}, the relationship \textit{born in} expects an entity of type \texttt{person} (\textit{Toriyama}) as input and \texttt{place} (\textit{Japan}) as output. We propose to decompose multi-hop questions into single-hop questions and iteratively resolve each relationship. If the question is decomposed correctly, the output types of one relationship are expected to overlap the input types of the next relationship. If there is no overlap between the input and output of neighboring relationships, then the semantic analysis has revealed an error that is required to be corrected before the relationship chain is evaluated to answer the MQA.

\section{Methodology}
\label{sec:methodology}
In this section, we present the methodology of the \name framework. The input to the \name framework is a multi-hop question and a set of factual edits. The output is an answer to the questions. The \name framework consists of a type extraction step, a multi-hop question decomposition step, and a subquestion resolution step. The flow of the framework is illustrated in Figure~\ref{fig:overview}. The type extraction step involves developing functions and a library for extracting the type of entities and relationships, respectively. The details of the type extraction is provided in Section~\ref{sec:type}.
%Entity types are extracted during MQA using an entity linking model and an LLM while relationship types are extracted prior to MQA to build a template library of relationship inputs and outputs, described in 
The multi-hop question decomposition step involves decomposing the multi-hop question into a chain of relationships that represent each hop and the initial multi-hop question entity. The relationship chain is then checked for type alignment and realigned if necessary. The details of question decomposition are explained in Section~\ref{sec:qa_decomp}. The final step is to iteratively traverse the relationship chain until the answer entity is found in the subquestion resolution step. Within each iteration, the entity and relationship are compared against edits to determine whether it is necessary to insert edited information. The details of the relationship chain traversal are provided in Section~\ref{sec:rc_traversal}.

\begin{figure*}
    \centering
    % \includesvg[width=\textwidth]{Figures/tc_example.svg}
    \includegraphics[width=\textwidth]{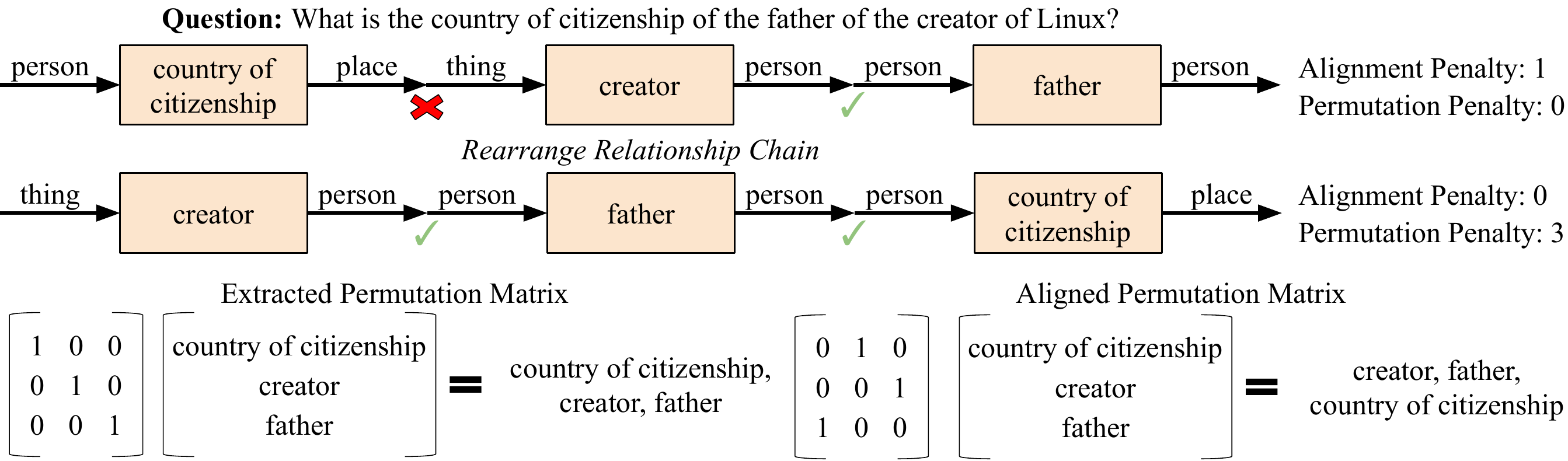}
    \caption{Example of relationship chain repair on a misaligned chain. Initially, there is an alignment penalty of 1. After two permutation steps, the relationship chain is realigned.}
    \label{fig:type_checking}
\end{figure*}

\subsection{Type Extraction}
\label{sec:type}
In this section, we describe how the types for both entities and relationships are extracted. The entity types are extracted using a combination of entity linking models and LLMs during MQA. In contrast, we pre-characterize a library of input and output types for the relationships. During MQA, the library is queried to obtain the input and output type of each relationship. The approach to generating types for entities and relationships is different because the number of different entities is very large and cannot be enumerated ahead of time. On the other hand, there is only a limited number of relationship that connect persons, places, and things. Therefore, it is possible to  pre-characterize the different types of relationships into a template library, for quick and reliable access at runtime.  

\textbf{Entity Type Extraction:} The objective of entity type extraction is to determine if an entity is of type person, place, or thing. We first pass the entity to an entity linking model, which can accurately decide whether the entity is a person or not. \name uses the ReFinED~\cite{refined} entity linking model for its short inference times and accurate entity linking. If the entity is not a person, then the entity is passed to the language model $\mathcal{F}$  to determine whether the entity is a place or thing. The prompt and in-context examples are provided in Section 7 of the Appendix.

\textbf{Relationship Type Extraction:} The relationships from given edits are extracted to build a relationship template library, where template relationships $r_t$ are encoded using a dense retrieval model and act as keys to access their input and output types. During MQA, the template relationships $r_t$ are compared against embedded multi-hop question relationships $r$ using cosine similarity. Question relationships $r$ take on the input and output types of the most similar $r_t$. Edits are expected to be provided as triples, so $r_t$ can be taken directly from the edits. Relationships can have multiple input and output types, but LLMs struggle to correctly assign multiple types at once. To generate an accurate template library, the relationship types are manually labeled, which is feasible due to the limited number of relationships. 

\begin{figure*}[t]
    \centering
    % \includesvg[width=\textwidth]{Figures/overview.svg}
    \includegraphics[width=\textwidth]{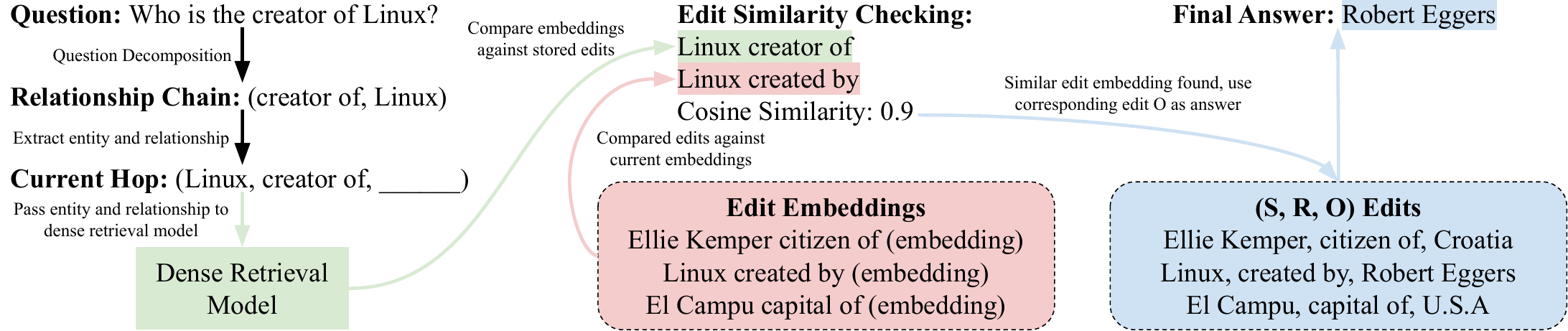}
    \caption{Example of \name answering the question \textit{Who was the creator of Linux?} The subject and relationship are extracted during question decomposition and rearranged into a single triple with the object unknown. The entity and relationship are embedded using a dense retrieval model and compared against the stored edit embeddings. A similar edit is found, so the object of the corresponding triple is used as the next entity / final answer.}
    \label{fig:example}
\end{figure*}

\subsection{Question Decomposition}
\label{sec:qa_decomp}

In this section, we explain how a relationship chain is extracted from a multi-hop question. \name receives the multi-hop question and extracts a chain of hops that will be traversed to find the answer to the multi-hop question. Then, the input and output entity types of each relationship in the chain are checked to ensure the types are aligned. If the chain is misaligned, it is reconstructed to create a new relationship chain that is aligned.

\textbf{Chain Extraction:} The multi-hop question $\mathcal{Q}$ is passed to the LLM $\mathcal{F}$ along with an in-context learning prompt found, in Section 7 of the Appendix, in order to extract the relationship chain $\mathcal{R}$. Relationship chain extraction is outlined as follows:
\begin{equation}
    \mathcal{R} = (r_0, r_1, ..., r_n) = \mathcal{F}(\mathcal{Q}),
\end{equation}
where $\mathcal{R}$ is a chain of relationships $(r_0, r_1, ..., r_n)$. The initial entity $o_n$ from $\mathcal{Q}$ is extracted using an entity linking model. The relationship chain $\mathcal{R}$ can be iteratively traversed backwards to generate triples $(o_n, r_n, o_{n-1})$ until $o_{n-1}$ is the final answer $o_0$ to $\mathcal{Q}$. The traversal process can be viewed as a series of function calls where $r_n$ is a function that takes $o_n$ as input and outputs $o_{n-1}$ such that $o_0 = r_1(r_2(r_n(o_n)))$. Relationship chain traversal is described in-depth in Section~\ref{sec:rc_traversal}. 

\textbf{Chain Alignment:} The extracted relationship chains $\mathcal{R}$ have been observed to contain misaligned relationships $r$. Misalignment occurs when the ordering of $r$ within $\mathcal{R}$ does not match the ordering of relationships within the original multi-hop question. To check the alignment of the extracted chain, all $r$ in $\mathcal{R}$ are given input $\mathcal{T}_{in}$ and output $\mathcal{T}_{out}$ types, as described in Section~\ref{sec:type}. The relationships types are combined to form a chain $\mathcal{C}$ of types corresponding to $\mathcal{R}$, such that:
\begin{equation}
    \mathcal{C} = [(\mathcal{T}^{0}_{in}, \mathcal{T}^{0}_{out}), (\mathcal{T}^{1}_{in}, \mathcal{T}^{1}_{out}), ..., (\mathcal{T}^{n}_{in}, \mathcal{T}^{n}_{out})].
\end{equation}

The alignment penalty $\mathcal{A}$ for a type chain $\mathcal{C}$ is determined by the number of input and output type pairs $n$ that are misaligned. The alignment of $\mathcal{C}$ is described as follows:
\begin{equation}
\label{eq:t_cost}
\begin{aligned}
    &\mathcal{A} = \sum_{i=1}^{n}m_i,\\
    &\text{s.t. } m_i = \begin{cases}
        0 & \mathcal{T}_{i}^{out} == \mathcal{T}_{i+1}^{in} \\
        1 & \mathcal{T}_{i}^{out} \neq \mathcal{T}_{i+1}^{in},
    \end{cases} 
\end{aligned}
\end{equation}
where a misalignment $\mathcal{T}_{i}^{out} \neq \mathcal{T}_{i+1}^{in}$ carries a penalty of $1$, while an alignment $\mathcal{T}_{i}^{out} == \mathcal{T}_{i+1}^{in}$ carries no penalty. The penalty values $m_i$ are summed to get $\mathcal{A}$. If $\mathcal{A} == 0$, $\mathcal{C}$ is properly aligned and is passed to the subquestion resolution step. If $\mathcal{A} > 0$, \name attempts to find an aligned $\mathcal{C}$.

\textbf{Relationship Chain Repair:} When a relationship chain $\mathcal{R}$ is not aligned, we have observed that $\mathcal{R}$ often contains the correct relationships, only in an incorrect order. Therefore, there exists an opportunity to correct the misaligned chains by permuting the relationships to find an aligned chain. The relationship chain repair step starts by generating all permutations of the relationship type chain $\mathcal{C}$. The repair is successful if one of the permutations $c$ has an alignment penalty of zero $\mathcal{A}(c) = 0$. However, there may be multiple permuted chains that have an alignment penalty of zero. If multiple $c$ exist where $A(c)=0$, we select $c$ that has the smallest permutation cost $\lambda$, the $c$ that required the smallest number of permutations  to eliminate the alignment penalty.
% Ideally, at least one permutation $c$ of $\mathcal{C}$ is fully aligned, so the objective of relationship chain repair is to find an aligned $c$. 

To find $\lambda$ for each $c$, the permutation matrix $\mathcal{P}$ used to generate $c$ is examined. Generating $c$ using $\mathcal{P}$ can be described as $\mathcal{P}\cdot\mathcal{R}=c$, and expanded as:
% The permutation cost $\lambda$ Each $c$ can be viewed as an input / output pair $c_{n}$ vector. A permutation matrix $\mathcal{P}$ is used to generate all $c$, described as $\mathcal{P}\times\mathcal{R}=c$ and expanded in:
\begin{equation}
    \begin{pmatrix}
        1 & 0 & ... & 0\\
        0 & 1 & ... & 0\\
        ... & ... & ... & ... \\
        0 & 0 & ... & 1\\
    \end{pmatrix}
    \begin{pmatrix}
        \mathcal{C}_0 \\
        \mathcal{C}_1 \\
        ... \\
        \mathcal{C}_n \\
    \end{pmatrix}
    = 
    \begin{pmatrix}
    c_0 \\
    c_1 \\
    ... \\
    c_n \\
    \end{pmatrix},
\end{equation}
where $\mathcal{C}_n$ and $c_n$ correspond to the input / output type pairs within the the original type chain $\mathcal{C}$ and its permutations $c$.

The rows in $\mathcal{P}$ can be reordered to obtain a different permutation $c$, as long as each row and column has exactly $1$ non-zero value. \name attempts to find a $c$ where $\mathcal{A}(c)=0$ and the permutation penalty $\lambda$ of $\mathcal{P}$ is minimal. The penalty $\lambda$ of the permutation matrix $\mathcal{P}$ is computed  as follows:
\begin{equation}
\begin{aligned}
\label{eq:permute_cost}
    \lambda = N - \sum_{i=1}^{N} \mathcal{P}_{i,i}, 
\end{aligned}
\end{equation}
where $\mathcal{P}_{r,c}$ denotes the element in row $r$ and column $c$ of $\mathcal{P}$. 
The summation computes the number of elements that are on the diagonal, which corresponds to the number of elements in their "unpermuted" position. Subtracting the sum from the number of diagonal elements $N$ provides the number of rows that have been permuted, also known as the permutation cost $\lambda$ for the current permutation $c$. The initial permutation matrix $\mathcal{P}$ will have the value $1$ on all diagonal matrix elements $\mathcal{P}_{i,i}$, so the $\lambda$ for $\mathcal{C}$ will be $0$, while $\lambda$ for any permutation $c$ will be non-zero.

Equation~\ref{eq:t_cost} is applied to all generated $c$. If at least $1$ permutation $c$ where $\mathcal{A}(c)=0$ exists, the $c$ where $A(c)=0$ with the lowest $\lambda$ is used as the new $\mathcal{C}$ and the corresponding permutation of $\mathcal{R}$ is passed to the subquestion resolution step. 

If an aligned $c$ does not exist, then question decomposition is restarted and a higher LLM temperature is used during chain extraction. We use a temperature scale of $0.0$ to $1.0$ on increments of $0.1$. After $1.0$, the model responses tend to become too varied to be useful. The $0.1$ increment gives a good trade-off between exploring many options and while still having a reasonable run time. If no aligned $c$ is found, the $c$ with the least $\mathcal{A}$ is used during subquestion resolution.

\input{Tables/mquake_acc}
\input{Tables/vs_older}

\subsection{Subquestion Resolution}
\label{sec:rc_traversal}
The input to the subquestion resolution step is an entity $o_n$ and relationship $r$. The goal is to generate the next entity $o_{n-1}$. Starting with the initial entity and relationship, the current $o_n$ and $r$ are compared against stored edits. If an edit that is sufficiently similar to one of the inputs exists, the next entity $o_{n-1}$ will be an edited object $o'$ from the corresponding edit triple $t'$. Otherwise, the LLM is prompted to generate and answer a question based on the current $o_n$ and $r$ to find $o_{n-1}$. This process is repeated until the final multi-hop answer has been obtained. Each entity and relationship is embedded using a dense retrieval model and is compared against all edit embeddings using cosine similarity. An example of subquestion resolution is provided in Figure~\ref{fig:example}.

\textbf{Edit Storage:} Prior to MQA, \name receives edits as triples and stores them as strings and embeddings. Edit triples $t'$ are stored as comma-separated lists. The subject $s$ of $t'$ is passed to an entity linking model  to get the true name $s^*$ of the entity. The edit triple $t'$ is then updated with $s^*$ such that $t'=(s^*, r, o')$. 

Edits $t'$ are also stored in an embedding space. The subject $s$ and relationship $r$ of $t'$ are combined into one string and passed to an embedding model  to get an edit embedding. Following previous works, we use the Contriever~\cite{contriever} dense retrieval model. These embeddings are then used during relationship chain traversal to aid in determining whether an edit needs to be made. 

\textbf{Edit Retrieval:} The initial entity $o_n$ is passed to an entity linking model  to generate its true name $o^{*}_{n}$. The initial entity true name $o^{*}_{n}$ is compared against all true edit names $s^*$ previously inserted into \nameNoSpace. Next, the list of edit triples $\mathcal{L}_t$ is searched for corresponding edits as follows:
\begin{equation}
    \mathcal{E}_{search} = 
    \begin{cases}
        \mathcal{E}_{s^{*}}, & \text{if } o^{*}_{n} \in \mathcal{L}_t, \\
        \mathcal{E}, & \text{if } o^{*}_{n} \notin \mathcal{L}_t,
    \end{cases}
\end{equation}
where the set of semantic embeddings that will be checked for necessary edits $\mathcal{E}_{search}$ is narrowed to the embeddings $\mathcal{E}_{s^{*}}$ where $s^*\in\mathcal{L}$ and $o^{*}_{n} == s^*$. The entity linking model  can generate false positive and false negative outputs, so even if no match is found, all edited semantic embeddings $\varepsilon_e \in \mathcal{E}$ are checked to ensure that an edit is not missed.

Once $\mathcal{E}_{search}$ has been found, the initial entity $o_n$ and relationship $r_n$ of $\mathcal{Q}$ are extracted from $\mathcal{R}$ and passed to a dense retrieval model  to generate semantic embeddings $\varepsilon_c$ for the current hop. Then, $\varepsilon_e \in \mathcal{E}$ are compared against $\varepsilon_c$ using cosine similarity $cos()$. If the highest similarity embedding is above a threshold $\tau$, then the $o'$ from the corresponding edit triple $t'$ becomes the new $o_{n-1}$. If the highest similarity is below $\tau$, then $o_n$ and $r_n$ move to the triple completion sub-step. Semantic embedding matching can be described algorithmically as follows:
\begin{equation}
    o_{n-1} = 
    \begin{cases}
        o_{e}, & \text{if } \text{cos}(\varepsilon_c, \varepsilon_e) > \tau, \\
        \text{None}, & \text{if } \text{cos}(\varepsilon_c, \varepsilon_e) <= \tau.
    \end{cases}
\end{equation}

If no new $o_{n-1}$ is found through semantic embedding similarity, then the LLM $\mathcal{F}$ is prompted to generate the next $o_{n-1}$. First, $\mathcal{F}$ is prompted using in-context learning to generate a question $\mathcal{Q}_{LLM}$ based on $o_n$ and $r_n$. The question-generating in-context learning prompt is provided in Section 7 of the Appendix. The LLM-generated question $\mathcal{Q}_{LLM}$ is then answered by $\mathcal{F}$ using another in-context learning prompt to ensure that only a single entity $o_{n-1}$ is provided as an answer.% The final in-context learning prompt is provided in Figure~\ref{fig:answer_icl_prompt}. 

Once $o_{n-1}$ is generated through one of the previous sub-steps, it is paired with $r_{n-1}$ to complete the previous two sub-steps to find $o_{n-2}$. This process is iteratively completed $[(o_n, r_n, o_{n-1})\rightarrow(o_{n-1}, r_{n-1}, o_{n-2})\rightarrow...\rightarrow(o_1, r_1, o_0)]$ until all $r_n$ have been used and the final answer $o_0$ is found.

\section{Results}
\label{sec:results}
The code for CHECK is available at \url{https://github.com/dominic-simon/CHECK}.

\textbf{Baselines:} We compare against other editors that do not rely on outside sources of factual information. Specifically, we compare against MeLLo~\cite{mquake}, PokeMQA~\cite{pokemqa}, DeepEdit~\cite{deepedit}, and the question-answering portion of GMeLLo~\cite{gmello}. We also provide comparisons against the parameter-based knowledge editors ROME~\cite{rome} and MEMIT~\cite{memit} as well as model finetuning. 

\textbf{Datasets:} We use the MQuAKE~\cite{mquake} dataset to evaluate the editors. MQuAKE is composed of two subsets. The counterfactual subset contains $3000$ edit cases. The subset contains questions with $2$, $3$, and $4$ hops with $1000$ cases of each. Each edit case contains betwen $1$ and $4$ individual edits. The temporal subset is composed of $1868$ edit cases containing $1421$ 2-hop questions, $445$ 3-hop questions, and $2$ 4-hop questions each with only $1$ edit. Two additional subsets have also been added to MQuAKE~\cite{deepedit}. The counterfactual subset contains conflicting edit cases, so MQuAKE-$2002$ removes all cases with conflicting edits, resulting in a counterfactual dataset containing only $2002$ edit cases. The other new subset contains $429$ edit cases each with $4$ hops and $4$ edits.

\textbf{Evaluation Metrics:} Each edit case in MQuAKE contains $3$ multi-hop questions conveying the same idea with the same number of hops in slightly different words. An edit case is considered correct if the editor correctly answers at least $1$ question. We also track the number of questions each editor has answered correctly. Per-case accuracy is determined as correct cases $\div$ total cases and per-question accuracy is determined as correct questions $\div$ total questions.

\textbf{Models:} We compare the baselines across $3$ models: GPT-J~\cite{gptj}, Vicuna-7B~\cite{vicuna}, and Falcon-7B~\cite{falcon7b}.

\textbf{Implementation:} MeLLo and PokeMQA were limited to $5$ hops to keep the experiment time reasonable. Similarly, DeepEdit was allowed $5$ additional knowledge candidates. Additionally, they were each allowed a maximum of $200$ new tokens to be generated for each LLM call. \name used a cosine similarity threshold of $0.8$ and was limited to a maximum of $50$ new tokens per model call.

\textbf{Hardware Setup:} All experiments were conducted on $1$ NVIDIA A100 GPU and $8$ CPU cores.

We present two core experiments in the following sections: an evaluation of \name against other knowledge editors across $4$ datasets and $3$ LLMs in Section~\ref{sec:main_acc}, and an ablation study on the performance of \name over varying numbers of hops and edits in  Section~\ref{sec:hop_edit_acc}. Additional experiments are provided in the Appendix. 

\subsection{Editing Accuracy}
\label{sec:main_acc}
The evaluation of \name and other state-of-the-art multi-hop knowledge editors is provided in Table~\ref{tab:mquake_acc}. Across all models, GMeLLo, MeLLo, DeepEdit, and PokeMQA consistently struggle to achieve a $20\%$ per-case accuracy on three of the four subsets. The only subset they are able to find better performance on is MQuAKE-T, which contains the least number of hops and edits per question. This is unsurprising as they all rely on LLMs doing large amounts of reasoning at once. DeepEdit is able to break the $20\%$ per-case accuracy mark in a few of the results and PokeMQA overpeforms on Vicuna when compared to its performance on GPT-J and Falcon. \name does not share this struggle of breaking $20\%$ per-case accuracy, achieving a 31.57\%, 28.51\%, 24.79\%, and 16.77\% increase in accuracy over the next highest on the MQuAKE-CF-3k, MQuAKE-2002, MQuAKE-Hard, and MQuAKE-T subsets respectively.  

Parameter-based knowledge editors have proven unsuccessful on mutli-hop questions. An evaluation of the accuracy of \name and other parameter-based knowledge editors on MQuAKE subsets is provided in Table~\ref{tab:vs_older}. The accuracies are from \cite{rae}. \name outperforming the parameter-based editors is in line with previous works, further affirming that storage-based knowledge editors are better able to handle the intermediate reasoning steps required for MQA than parameter-based knowledge editors.

\subsection{MQA Hop and Edit Ablation}
\label{sec:hop_edit_acc}

\begin{figure}[t]
    \centering
    \includegraphics[width=0.85\linewidth]{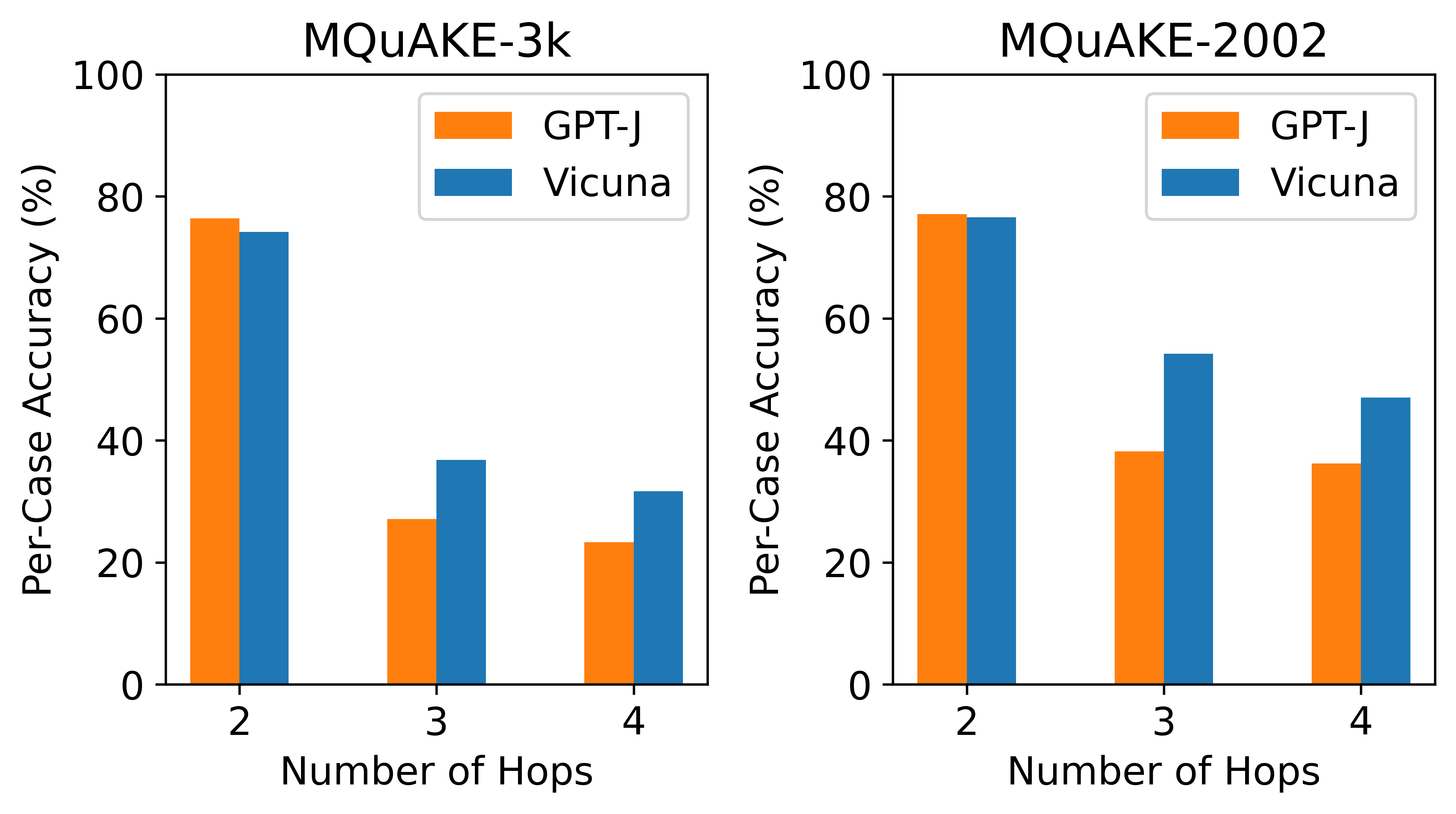}
    \caption{The accuracy of \name on the MQuAKE-3k and MQuAKE-2002 datasets across different numbers of question hops.}
    \label{fig:perhop_acc}
\end{figure}
\begin{figure}[t]
 \centering   \includegraphics[width=0.85\linewidth]{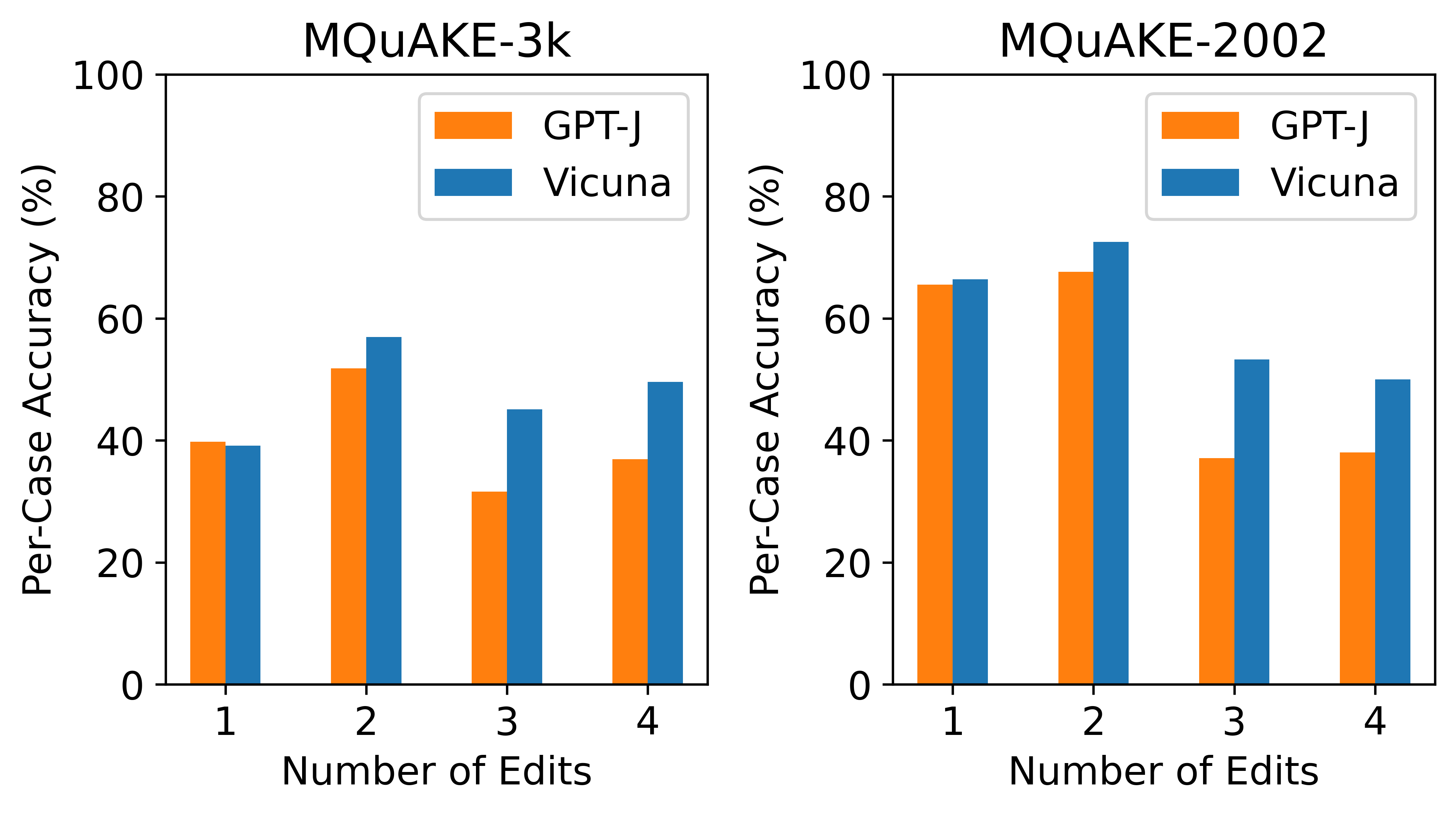}
    \caption{The accuracy of \name on the MQuAKE-3k and MQuAKE-2002 datasets across different numbers of edits per multi-hop question.}
    \label{fig:peredit_acc}
\end{figure}

We provide a breakdown of \nameNoSpace's per-case accuracy on MQuAKE-3k and MQuAKE-2002 over the number of hops and edits in Figure~\ref{fig:perhop_acc} and Figure~\ref{fig:peredit_acc}, respectively. As the the number of hops increases, the accuracy of \name decreases. This is an expected outcome since longer multi-hop questions require longer relationship chains, introducing more areas for both question decomposition and subquestion resolution to fail. Similarly, as the number of edits increases, the accuracy also drops. Greater number of edits correspond to longer questions, which are more difficult to correctly answer. \nameNoSpace's over-performance on questions with 2 edits can be attributed to the in-context learning prompt used during question decomposition.

\section{Conclusion}
\label{sec:conclusion}
We present the \name framework for multi-hop knowledge editing. The main insight of \name is that the LLM subquestion reasoning process can be checked for semantic consistency. \name decomposes multi-hop questions into a chain of relationships and ensures the semantic consistency of that chain. The chain is then iteratively traversed, answering each of the subquestions that make up the chain and inserting edits where necessary until the answer to the multi-hop question is reached. %Using this approach, \name achieves a $22.8$\% average increase in accuracy over similar knowledge editors. 

\newpage

\section*{Acknowledgements}
This material is in part sponsored by UF startup funds and DARPA under agreement number FA8750-23-2-0501. The views and conclusions contained herein are those of the authors and should not be interpreted as necessarily representing the official policies or endorsements, either expressed or implied, of DARPA or the U.S. Government. 

\bibliographystyle{named}
\bibliography{ijcai24}

\newpage

\appendix

\section{Overview}
An additional 6 sections are provided in this Appendix. The reasoning of why the relationship chain is extracted in a reverse order is explained in Section~\ref{sec:rc_ordering}. %Results on \nameNoSpace's accuracy broken down by number of edits and number of hops per questions is discussed in Section~\ref{sec:hop_edit_acc}. 
An additional experiment on the ability of LLMs to correctly extract relationships from a multi-hop question is provided in Section~\ref{sec:rc_correct}. Additional experiments analyzing the effectiveness of regenerating relationship chains at higher LLM temperatures are provided in Section~\ref{sec:temp_per_hop}. More insight into \nameNoSpace's improved method of handling edits is given in Section~\ref{sec:cos_vs_dot}. An ablation study on the threshold value used during subquestion resolution is provided in Section~\ref{sec:threshold_ablation}. Finally, the in-context learning prompts used by \name are displayed in Section~\ref{sec:icl_prompts}.

\section{Relationship Chain Ordering}
\label{sec:rc_ordering}

\begin{figure}[t]
    \centering
    \includegraphics[width=0.75 \linewidth]{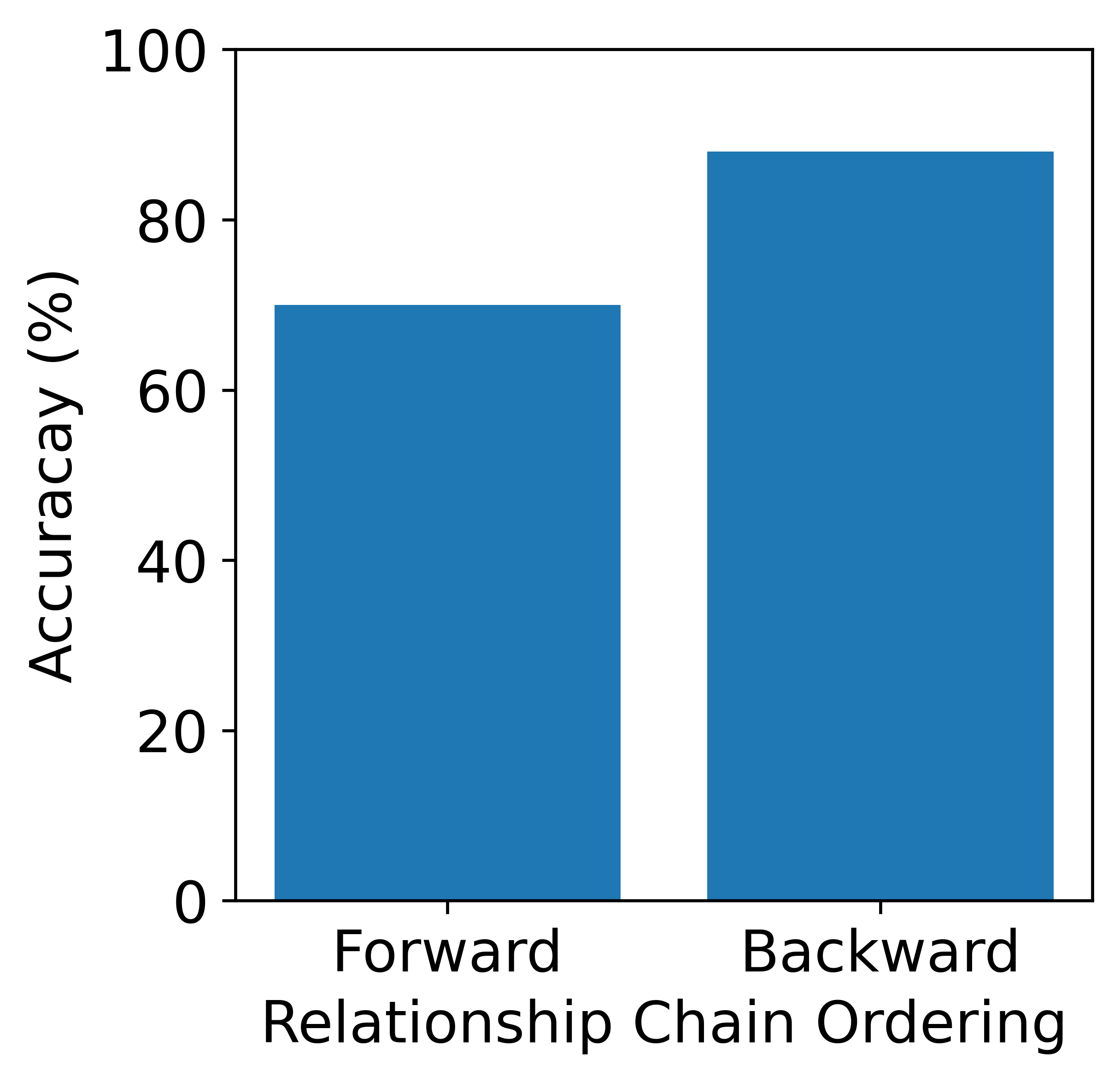}
    \caption{Accuracy of the \name\ with GPT-J when extracting a forward and backward relationship chain $\mathcal{R}$ over 300 multi-hop questions.}
    \label{fig:f_v_b}
\end{figure}

\begin{figure}[t]
    \centering
    \includegraphics[width=\linewidth]{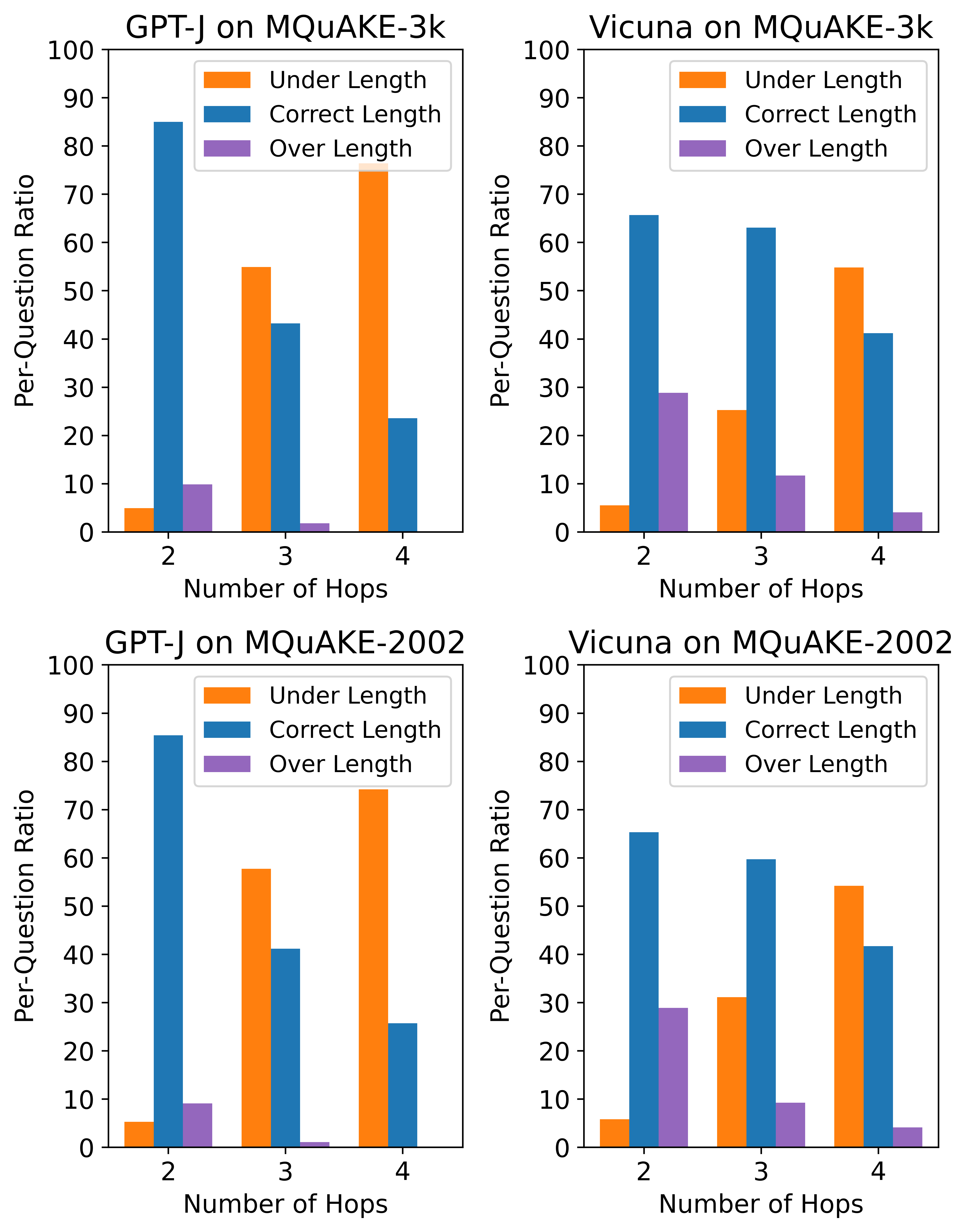}
    \caption{Analysis of the correctness of the lengths of relationship chains extracted during question decompositon by \name on MQuAKE-3k and MQuAKE-2002.}
    \label{fig:rc_correct}
\end{figure}

A more human intuitive ordering of $\mathcal{R}$ would be $(o_n, r_n, r_{n-1}, ..., r_0)$, where the initial entity $o_n$ is provided first with the initial relationship $r_n$ so that $\mathcal{R}$ can be traversed in forward order. However, a forward ordering of $\mathcal{R}$ is the opposite of how tokens are attended to in the masked attention components of $\mathcal{F}$. This causes $\mathcal{F}$ to perform additional reasoning to correct the relationship chain to the desired order. The current ordering of $\mathcal{R}$ directly reflects the direction in which the tokens of $\mathcal{Q}$ are attended to within $\mathcal{F}$. The ordering of $\mathcal{R}$ reflecting the masked attention mechanism in $\mathcal{F}$ allows $\mathcal{F}$ to extract a more accurate $\mathcal{R}$. We provide the accuracy of the full \name\ knowledge editor with GPT-J~\cite{gptj} when using a forward and reversed $\mathcal{R}$ over $300$ multi-hop questions in Figure~\ref{fig:f_v_b}. Extracting $\mathcal{R}$ in reverse order results in extracted relationships more closely resembling those in the question, resulting in higher knowledge editing accuracy.

\begin{figure}[t]
    \centering
    \includegraphics[width=\linewidth]{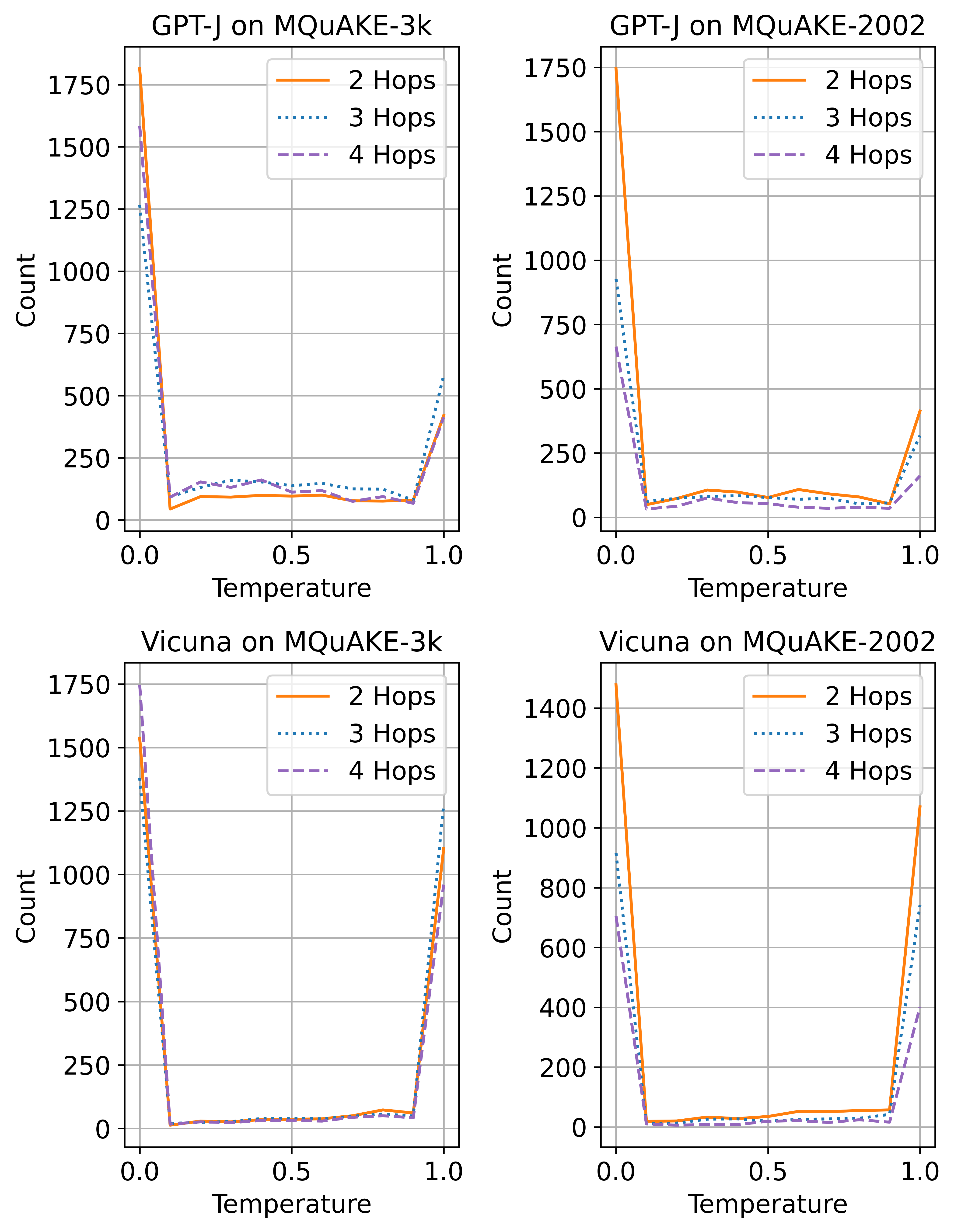}
    \caption{The LLM temperature at which the relationship chains found during question decomposition were generated at, categorized by the number of hops in the multi-hop question.}
    \label{fig:total_temp_hop}
\end{figure}

\section{Relationship Chain Correctness}
\label{sec:rc_correct}

The lengths of relationship chains extracted during \nameNoSpace's question decomposition step on the MQuAKE-3k and MQuAKE-2002 datasets are provided in Figure~\ref{fig:rc_correct}. The lengths are broken into three categories: under length chains contained less than the true number of relationships, correct length chains contained the true number of relationships, and over length chains contained more than the true number of relationships present in the multi-hop question. Each hop corresponds to a single relationship. For a 3-hop question, a relationship chain with 2 relationships is under length, a relationship chain with 3 relationships is correct length, and a relationship chain with 4 relationships is over length. 

Across both datasets and models, the number of correct length chains decreases as the number of hops increases. Conversely, the number of under length chains increases as the number of hops increases. These results parallel the results of Figure~\ref{fig:perhop_acc}, where \nameNoSpace's accuracy decreased as the number of hops increased. The most likely cause for this decrease in correctly extracted relationship chains is that the number of given examples in the in-context learning prompt for relationship chain extraction is small. There are 4 examples: 2 2-hop examples, 1 3-hop example, and 1 4-hop example. The set of examples likely skews the LLM towards extracts fewer relationships from the multi-hop question, resulting in more under length chains and lower MQA accuracy. We believe this results can be improved by including more examples with a greater question hop diversity in the in-context learning prompt. 

\section{Relationship Chain Temperatures}
\label{sec:temp_per_hop}

\begin{figure}[t]
    \centering
    \includegraphics[width=\linewidth]{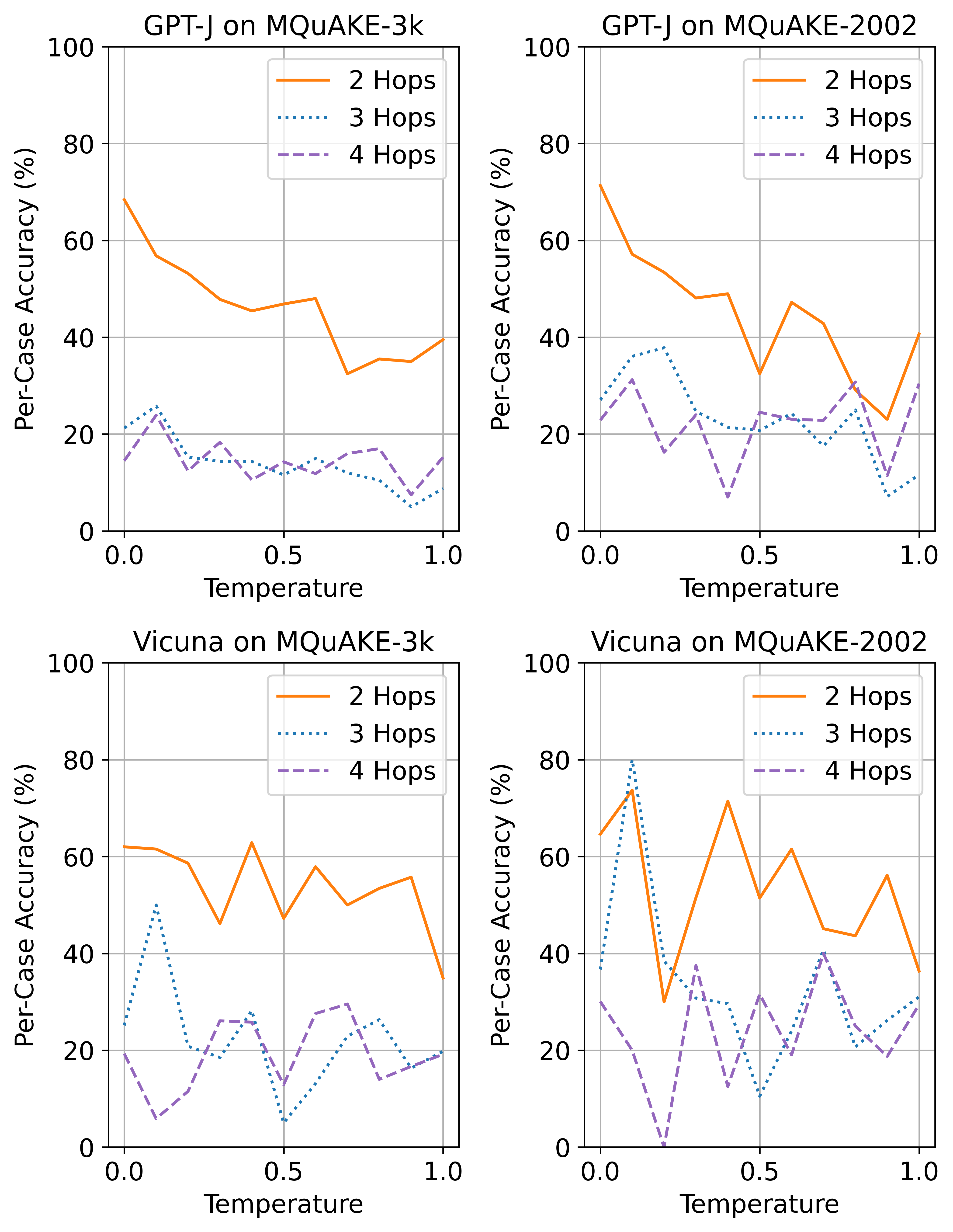}
    \caption{The accuracy of \name on the MQuAKE-3k and MQuAKE-2002 datasets separated by the temperature at which the final relationship chain was extracted. Each line represents the number of hops in the tested questions.}
    \label{fig:temp_hop}
\end{figure}

During the question decomposition step of \name, misaligned relationship chains are discarded and re-extracted at higher model temperatures in an attempt to generate an aligned relationship chain. The temperature of the relationship chain that gets used broken down by the number of hops is provided in Figure~\ref{fig:total_temp_hop}. Relationships chains that were still misaligned after being generated with a temperature of $1.0$ were added to the $1.0$ bin. Across both datasets and all models, the majority of relationship chains are correctly aligned on the first try, regardless of the hop count. As the temperature increases from $0.1$ to $0.9$, about 50 - 150 aligned relationships chains are generated and used for each of the temperature values. By generating misaligned relationship chains at higher temperatures, the chance to extract an aligned chain is increased. This allows \name to correctly answer questions that would not have been possible to correctly answer previously. There is then a spike in generated relationship chains again at the $1.0$ cutoff temperature value due to the multi-hop questions that \name was unable to extract an aligned chain for. 

\begin{figure*}[t]
    \centering
    \includegraphics[width=0.49\linewidth]{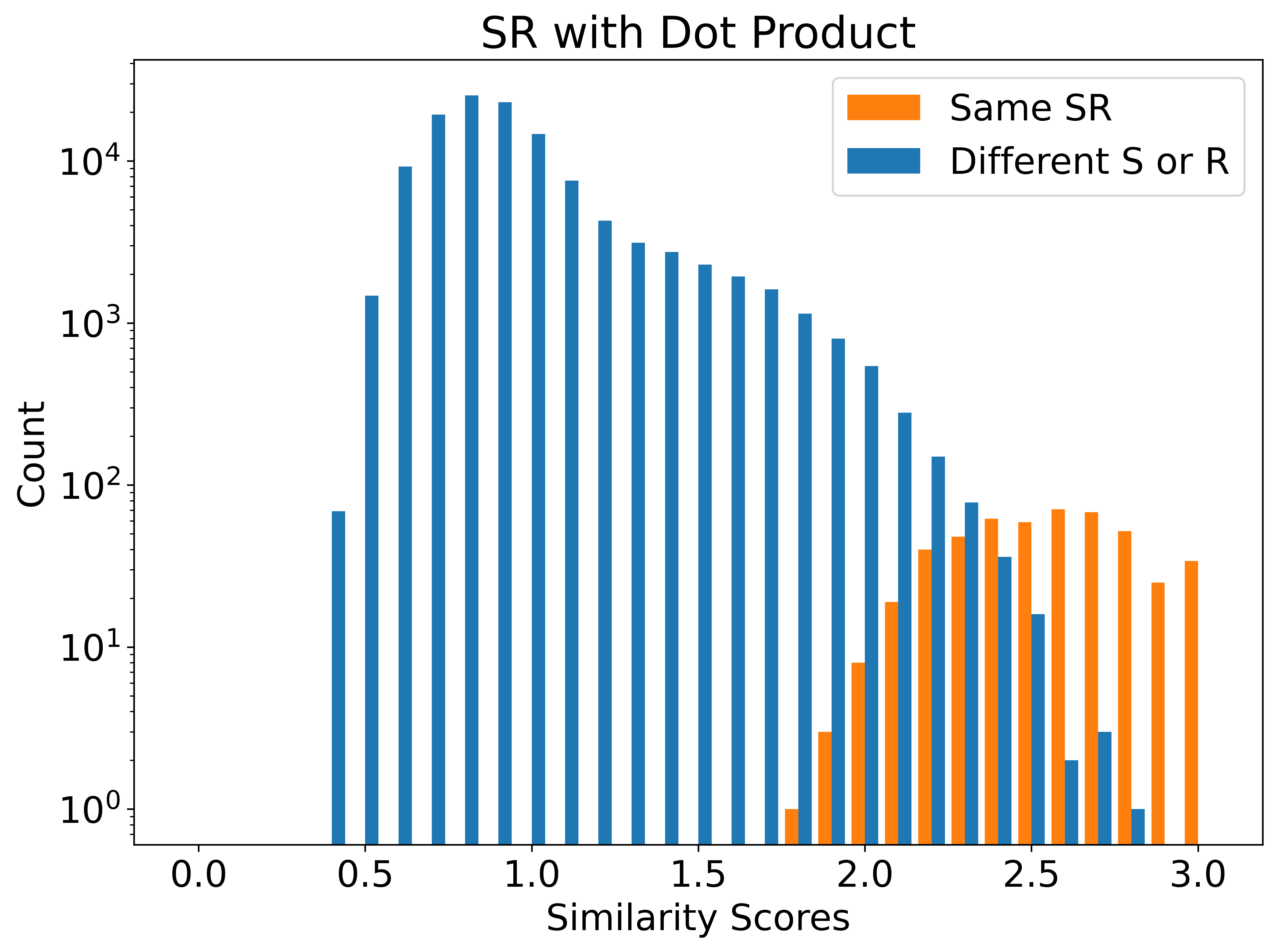} 
    \includegraphics[width=0.49\linewidth]{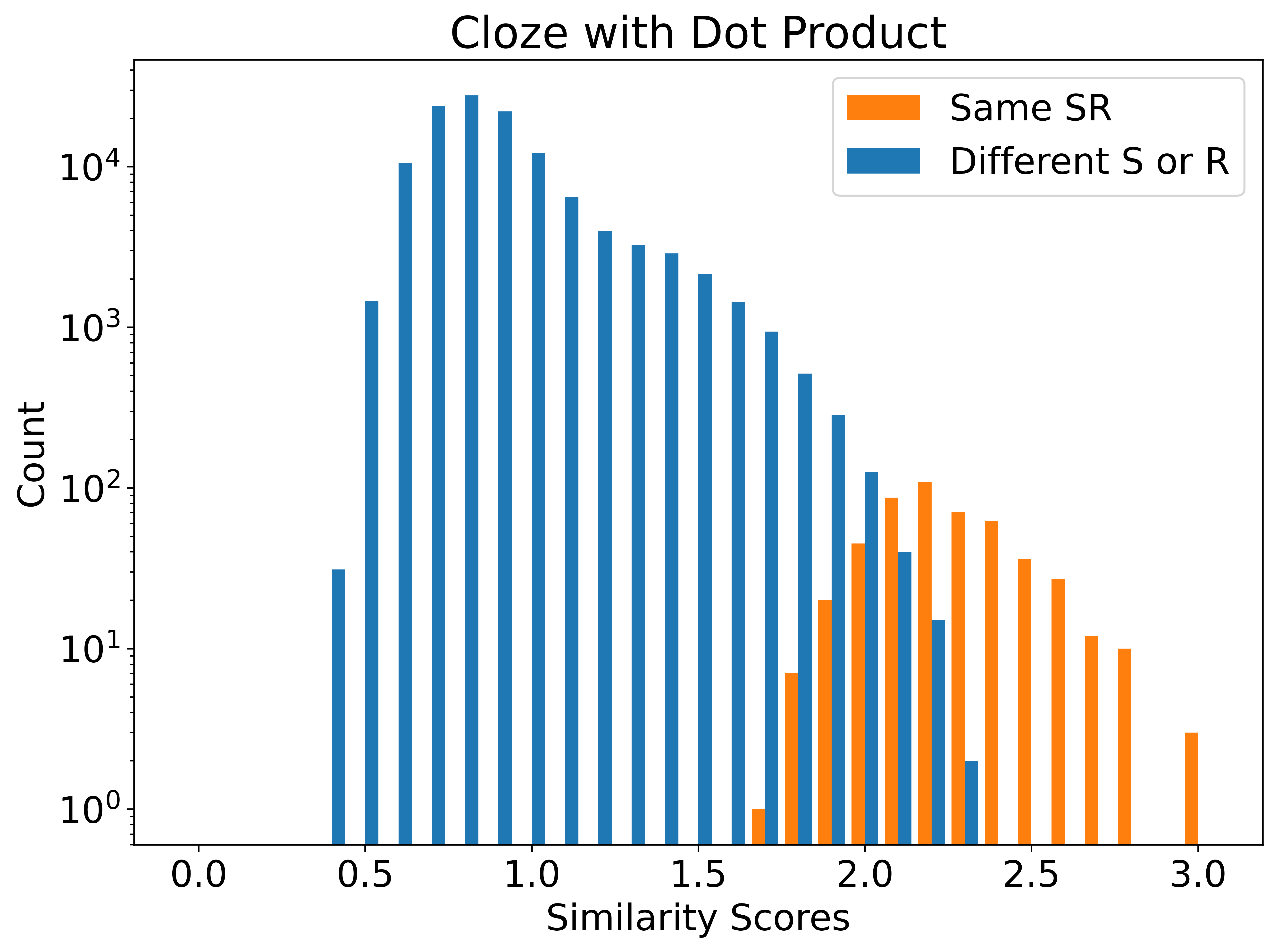} \\
    \includegraphics[width=0.49\linewidth]{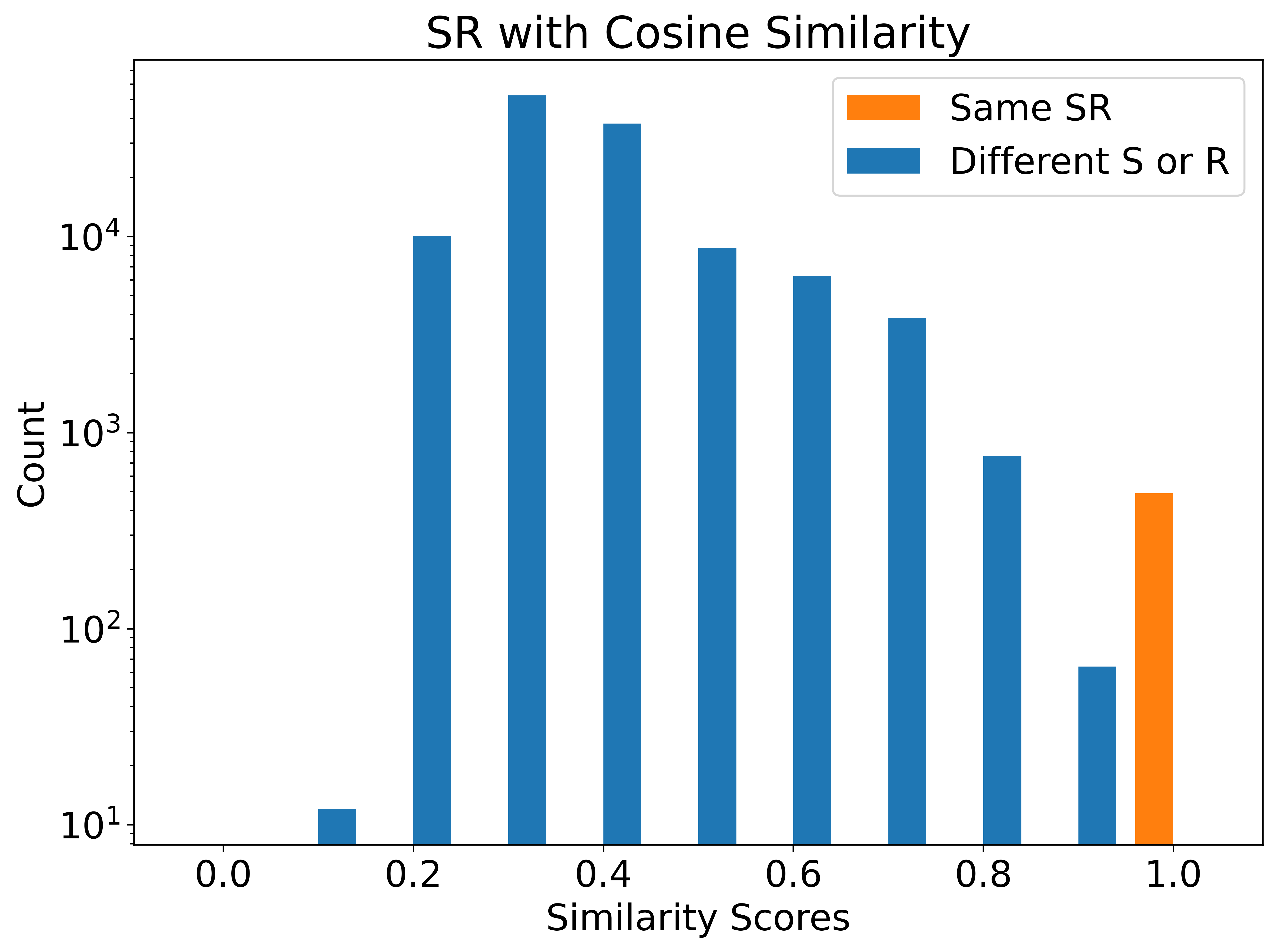}  
    \includegraphics[width=0.49\linewidth]{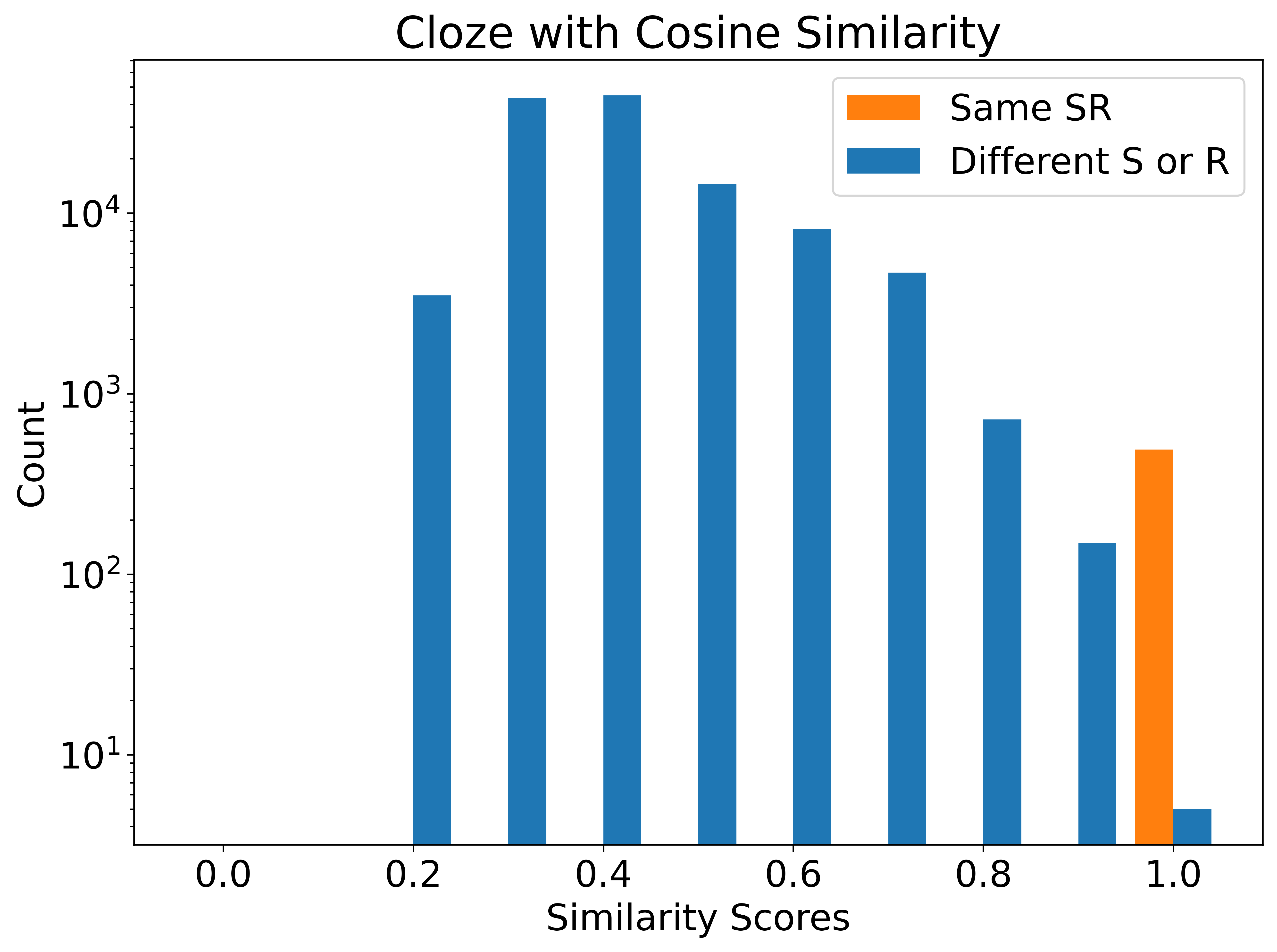} \\
    \caption{The similarity values between 500 embeddings. The embeddings style and similarity metric used are, from left to right, top to bottom: SR embedding with dot product similarty, cloze-style embedding with dot product similarity, SR embedding with cosine similarity, and close-style embedding with cosine similarity.}
    \label{fig:cos_vs_dot}
\end{figure*}

The accuracy of \name based on the LLM temperature at which the final relationship chain was generated is presented in Figure~\ref{fig:temp_hop}, broken by the number of multi-hop question hops. Across all graphs, the accuracy trends downward as the LLM temperature increases. However, having a non-zero accuracy in the temperatures greater than $0.0$ is a positive. Previously, these questions were incorrectly answered due to misalignment. Regenerating the relationship chains at higher temperatures has given \name the opportunity to correctly answer more questions.

\section{New Embedding Approach}
\label{sec:cos_vs_dot}

Many entity names are similar, for example \textit{John Adams} and \textit{John Hamm}. When embedded, it becomes difficult to separate entities with similar names from entities that are actually the same. Previous knowledge editors have have relied on dense retrieval models to embed edits and used dot product to determine the similarity between the current hop information and edited information. However, dot product produces a wide range of values that have a large overlap between similar and exact entity embeddings. Instead, we propose to only embed the subject and relationship of the edits and use cosine similarity to compare.

We provide the similarity values between $500$ embedded subject and relationship pairs, separated by the similarity metric used and the the the type of string that was embedded, in Figure~\ref{fig:cos_vs_dot}. There were $50$ different entities used, each with some variation of the name \textit{John} and a different last name. There were $10$ relationships used, chosen based on variety. The $50$ entities and $10$ relationships were combined to get the $500$ subject-relationship pairs. Each pair is embedded as the subject-relationship (SR) string and as a cloze-style sentence. SR embeddings are compared against SR strings only, and cloze embeddings are compared against cloze embeddings only. Each embedding is compared against both itself and the other embeddings. 

The left two graphs show the similarity values between the $500$ pairs when using dot product as the similarity metric. For both embedding types, there is a large amount of overlap between the values of similar embeddings and exact embeddings. This makes it difficult to discern a clear threshold that can be used to separate pairs that are similar and pairs that are the same. The right two graphs show the similarity values between the $500$ pairs when using cosine similarity as the similarity metric. There is a much clearer separation between similar and exact embedding similarity values. In fact, there is no overlap between similar and exact embedding similarity values. The small amount of similar embedding values in the 1.0 bin in the cloze cosine graph can be attributed to rounding. The SR embeddings have less similar embedding similarity values in the $0.9$ bin than the cloze embeddings, meaning that comparing SR embeddings with cosine similarity contains the clearest separation between exact and similar embeddings. For this reason, \name check embeds edits and current entity-relationship pairs as SR embeddings and uses cosine similarity to determine whether the current entity-relationship pair is similar to any edit.

\section{Threshold Ablation}
\label{sec:threshold_ablation}

\begin{figure}[t]
    \centering
    \includegraphics[width=\linewidth]{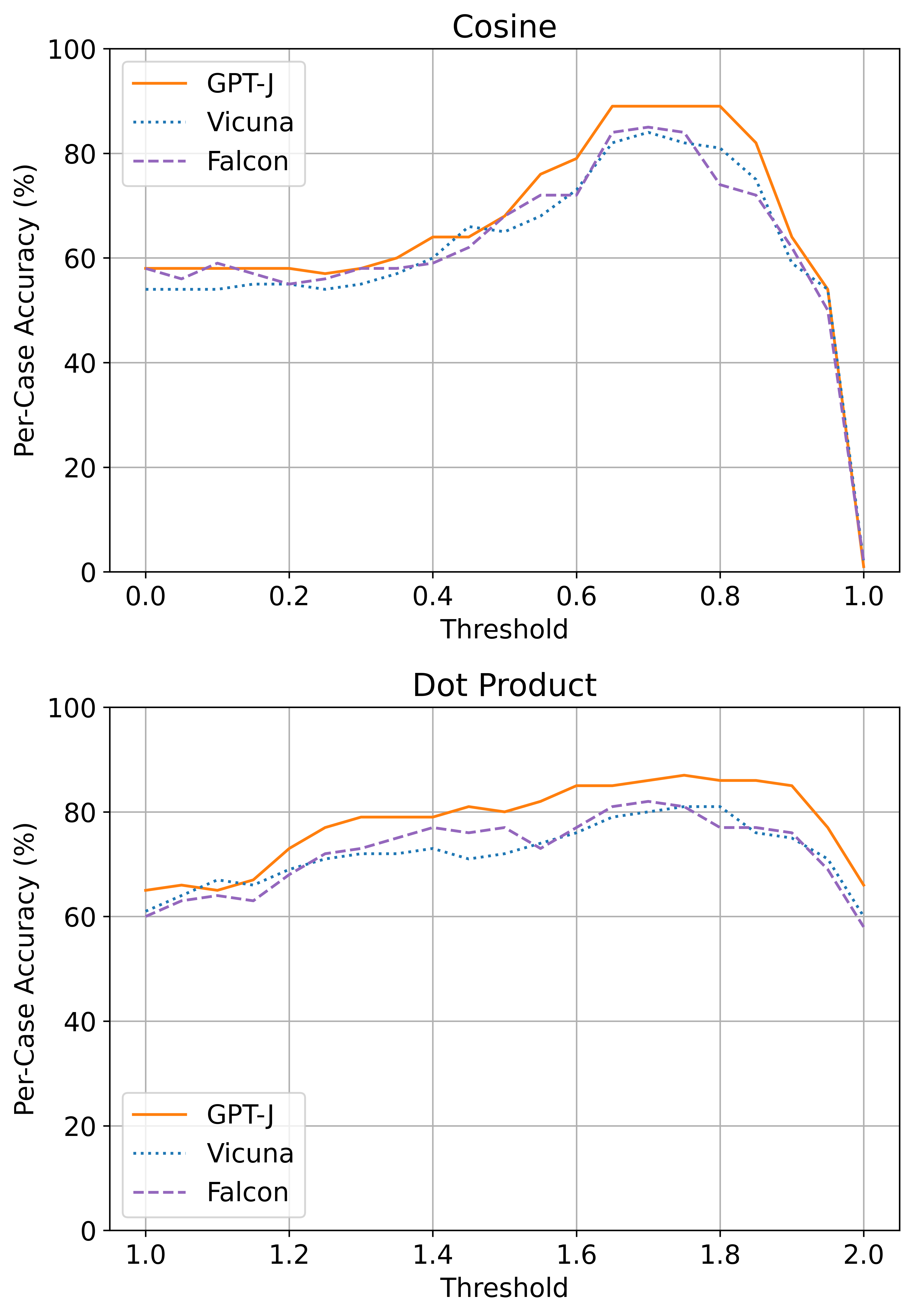}
    \caption{\nameNoSpace's accuracy at different similarity threshold values and functions.}
    \label{fig:thresh_ablate}
\end{figure}

We provide an ablation study on \name over the cosine and dot product similarity measures in Figure~\ref{fig:thresh_ablate}. The range for the cosine similarity results is $0.0$ to $1.0$ on $0.05$ increments. The range for the dot product similarity results is $1.0$ to $2.0$ on $0.05$ increments. Each increment is the accuracy of \name on the first 100 cases of MQuAKE-3k. For both cosine similarity and dot product, \name's accuracy steadily increases from thresholds of $0.0$ and $1.0$ to $0.75$ and $1.75$, respectively. The accuracy peaks between a threshold values of $0.75$ to $0.85$ and $1.75$ to $1.85$, before steeply decreasing from $0.90$ and $1.90$ to the end. The accuracies achieved using cosine similarity are higher than those achieved using dot product similarity. This is most apparent when \name is run with the Vicuna-7B LLM. \name uses the mean peak threshold of $0.80$ in all other experiments.

\section{In-context Learning Prompts}
\label{sec:icl_prompts}

This section contains the in-context learning prompts used by \name. The prompt used to extract relationships to form a relationship chain is in Figure~\ref{fig:icl_extract}. The prompt used to determine whether an entity is a place or a thing is given in Figure~\ref{fig:type_extract}. The in-context learning prompt used to generate a question from a given entity and relationship is provided in Figure~\ref{fig:qa_icl_prompt} and the prompt used to answer the generated question is in Figure~\ref{fig:answer_icl_prompt}.

\begin{figure}[t]
\noindent\fbox{%
    \parbox{\linewidth}{%
    Question: Who is the president of the country the Eiffel Tower is located in? \\
    SRO: $\vert$ president of $\vert$ country of $\vert$ \\ \\
    Question: What is the capital city of the country of citizenship of Ivanka Trump's spouse? \\
    SRO: $\vert$ capital city $\vert$ country of citizenship $\vert$ spouse of $\vert$ \\ \\
    Question Who is the spouse of the head of state in the United States of America? \\
    SRO: $\vert$ spouse of $\vert$ head of state $\vert$ \\ \\
    Question: On which continent is the country of citizenship of the founder of the manufacturer of iPhone 5 situated? \\
    SRO: $\vert$ continent of $\vert$ country of citizenship $\vert$ founder of $\vert$ manufacturer of $\vert$
        }%
}
    \caption{The in-context learning prompt provided to the LLM to extract $\mathcal{R}$}
    \label{fig:icl_extract}
\end{figure}

\begin{figure}[t]
\noindent\fbox{%
\parbox{\linewidth}{%
Choose between place, thing to describe the following entity:\\
Entity: Eiffel Tower\\
Type: thing\\

Choose between place, thing to describe the following entity:\\
Entity: Antarctica\\
Type: place\\

Choose between place, thing to describe the following entity:\\
Entity: soccer\\
Type: thing\\

Choose between place, thing to describe the following entity:\\
Entity: Japan\\
Type: place
}%
}
    \caption{The in-context learning prompt provided to the LLM the entity type $\mathcal{T}$ from the initial entity.
    }
    \label{fig:type_extract}
\end{figure}

\begin{figure}[t]
\noindent\fbox{%
    \parbox{\linewidth}{%
    Rephrase the following triple as a question: \\
    Triple: $\vert$ Eiffel Tower $\vert$ located in $\vert$ \\
    Question: What country is the Eiffel Tower located in? \\ \\
    Rephrase the following triple as a question: \\
    Triple: $\vert$ Joe Biden $\vert$ spouse of $\vert$ \\
    Question: Who is the spouse of Joe Biden? \\ \\
    Rephrase the following triple as a question: \\
    Triple: $\vert$ Ivanka Trump $\vert$ spouse of $\vert$ \\
    Question: What country is the Eiffel Tower located in? \\ \\
    Rephrase the following triple as a question: \\
    Triple: $\vert$ Rainn Wilson $\vert$ country of citizenship $\vert$ \\
    Question: What is the country of citizenship of Rainn Wilson? \\ \\
    Rephrase the following triple as a question: \\
    Triple: $\vert$ iPhone5 $\vert$ manufactured by $\vert$ \\
    Question: Who manufactures the iPhone5?
        }%
}
    \caption{The in-context learning prompt provided to the LLM to generate a question based on $o_n$ and $r_n$.}
    \label{fig:qa_icl_prompt}
\end{figure}

\begin{figure}[t]
\noindent\fbox{%
    \parbox{\linewidth}{%
    Question: What country is the Eiffel Tower located in? \\
    Answer: France \\ \\
    Question: Who is the spouse of Joe Biden? \\
    Answer: Jill Biden \\ \\
    Question: Who is the spouse of Ivanka Trump? \\
    Answer: Jared Kushner \\ \\
    What is the country of citizenship of Rainn Wilson? \\
    Answer: United States of America \\ \\
    Question Who manufactures the iPhone5? \\
    Answer: Apple
        }%
}
    \caption{The in-context learning prompt provided to the LLM to generate $o_{n-1}$ based on $\mathcal{Q}_{LLM}$.}
    \label{fig:answer_icl_prompt}
\end{figure}

\end{document}

%% file: Tables/program_synthesis.tex
\begin{table}[]
    \centering
    \begin{tabular}{c||cc}
    \midrule
         % & \multicolumn{2}{c}{} \\
    Method & Question & Verification of\\
    & Decomposition & Decomposition\\
    \midrule
    GMeLLo & $\checkmark$ & $\emptyset$  \\
    MeLLo & $\checkmark$ & $\emptyset$  \\
    DeepEdit & $\checkmark$ & $\emptyset$  \\
    PokeMQA & $\checkmark$ & $\emptyset$  \\ 
    \name & $\checkmark$ & $\checkmark$  \\
    \midrule
    \end{tabular}
    \caption{Knowledge Editing steps included by state-of-the-art frameworks~\protect\cite{gmello,mquake,deepedit,pokemqa}.}
    \label{tab:program_synthesis}
\end{table}

%% file: Tables/mquake_acc.tex
\begin{table*}[t]
    \centering
    \begin{tabular}{c||rrrrrrrr}
        \midrule
        Dataset~\cite{mquake} & \multicolumn{2}{c|}{MQuAKE-CF-3k} & \multicolumn{2}{c|}{MQuAKE-2002} & \multicolumn{2}{c|}{MQuAKE-Hard} & \multicolumn{2}{c}{MQuAKE-T}  \\
        \midrule
        \multicolumn{1}{c||}{Accuracy Type} & \multicolumn{1}{c|}{Case} & \multicolumn{1}{c|}{Question} & \multicolumn{1}{c|}{Case} & \multicolumn{1}{c|}{Question} & \multicolumn{1}{c|}{Case} & \multicolumn{1}{c|}{Question} & \multicolumn{1}{c|}{Case} & \multicolumn{1}{c}{Question} \\
        \midrule
        \multicolumn{1}{c||}{Model} & \multicolumn{7}{c}{GPT-J~\cite{gptj}} & Size: 6B \\
        \midrule
        \multicolumn{1}{l||}{GMeLLo-QA~\cite{gmello}} & 10.60 & 6.04 & 10.39 & 6.14 & 8.86 & 4.35 & 21.95 & 10.67 \\
        \multicolumn{1}{l||}{MeLLo~\cite{mquake}} & 14.97 & 6.89 & 17.18 & 8.13 & 6.76 & 2.64 & 32.82 & 18.49 \\
        \multicolumn{1}{l||}{DeepEdit~\cite{deepedit}} & 19.03 & 13.44 & 27.17 & 19.55 & 6.53 & 3.96 & 55.84 & 41.86 \\
        \multicolumn{1}{l||}{PokeMQA~\cite{pokemqa}} & 15.70 & 6.97 & 19.98 & 8.72 & 11.66 & 5.59 & 59.37 & 31.00 \\
        \multicolumn{1}{l||}{\name} & \textbf{42.27} & \textbf{29.57} & \textbf{56.59} & \textbf{40.86} & \textbf{35.90} & \textbf{23.85} & \textbf{78.69} & \textbf{55.82} \\
        \midrule
        \multicolumn{1}{c||}{Model} & \multicolumn{7}{c}{Vicuna~\cite{vicuna}} & Size: 7B \\
        \midrule
        \multicolumn{1}{l||}{GMeLLo-QA~\cite{gmello}} & 11.23 & 6.44 & 10.84 & 6.41 & 5.59 & 2.41 & 28.53 & 14.38 \\
        \multicolumn{1}{l||}{MeLLo~\cite{mquake}} & 9.93 & 5.08 & 9.84 & 5.13 & 1.86 & 0.85 & 68.52 & 50.18 \\
        \multicolumn{1}{l||}{DeepEdit~\cite{deepedit}} & 13.87 & 8.38 & 20.63 & 12.52 & 0.93 & 0.54 & 34.05 & 19.04 \\
        \multicolumn{1}{l||}{PokeMQA~\cite{pokemqa}} & 30.97 & 18.18 & 40.51 & 25.66 & 30.77 & 15.70 & 68.68 & 48.11 \\
        \multicolumn{1}{l||}{\name} & \textbf{47.57} & \textbf{30.93} & \textbf{63.74} & \textbf{41.99} & \textbf{48.72} & \textbf{29.68} & \textbf{81.64} & \textbf{55.84} \\
        % \multicolumn{1}{l||}{\name\ - Vicuna 13B} & \textbf{} & \textbf{} & \textbf{} & \textbf{} & \textbf{} & \textbf{} & \textbf{} & \textbf{} \\
        \midrule
        \multicolumn{1}{c||}{Model} & \multicolumn{7}{c}{Falcon~\cite{falcon7b}} & Size: 7B \\
        \midrule
        \multicolumn{1}{l||}{GMeLLo-QA~\cite{gmello}} & 7.77 & 4.27 & 6.50 & 3.63 & 5.36 & 3.34 & 16.38 & 7.57 \\
        \multicolumn{1}{l||}{MeLLo~\cite{mquake}} & 4.01 & 7.30 & 10.14 & 5.56 & 1.63 & 0.85 & 52.94 & 36.42 \\
        \multicolumn{1}{l||}{DeepEdit~\cite{deepedit}} & 13.37 & 8.23 & 19.53 & 12.02 & 2.80 & 1.24 & 59.85 & 45.38 \\
        \multicolumn{1}{l||}{PokeMQA~\cite{pokemqa}} & 15.77 & 7.64 & 19.93 & 9.14 & 13.05 & 7.46 & 63.97 & 37.76 \\
        \multicolumn{1}{l||}{\name} & \textbf{39.10} & \textbf{24.10} & \textbf{52.80} & \textbf{33.72} & \textbf{45.22} & \textbf{31.08} & \textbf{81.69} & \textbf{57.51} \\
        \midrule
    \end{tabular}
    \caption{Per-case and per-question accuracy across the MQuAKE subsets. The highest accuracy per column and per model is bolded. The second highest accuracy is underlined.}
    \label{tab:mquake_acc}
\end{table*}

%% file: Tables/vs_older.tex
\begin{table}[t]
    \centering
    \begin{tabular}{c||rr}
        \midrule
        Dataset & \multicolumn{1}{c|}{MQuAKE-CF-3k} & \multicolumn{1}{c}{MQuAKE-T}  \\
        % \midrule
        % \multicolumn{1}{c||}{Accuracy Type} & \multicolumn{2}{c}{Case} \\
        \midrule
        \multicolumn{1}{c||}{Model} & \multicolumn{1}{c}{GPT-J} & \multicolumn{1}{r}{Size: 6B}\\
        \midrule
        \multicolumn{1}{l||}{FT*} & 7.70 & 3.10 \\
        \multicolumn{1}{l||}{ROME*} & 7.60 & 4.10 \\
        \multicolumn{1}{l||}{MEMIT*} & 8.10 & 10.60 \\
        \multicolumn{1}{l||}{\name} & \textbf{42.27} & \textbf{78.69} \\
        \midrule
        \multicolumn{1}{c||}{Model} & \multicolumn{1}{c}{Vicuna} & \multicolumn{1}{r}{Size: 7B}\\
        \midrule
        \multicolumn{1}{l||}{FT*} & 4.80 & 23.10 \\
        \multicolumn{1}{l||}{ROME*} & 8.40 & 5.00 \\
        \multicolumn{1}{l||}{MEMIT*} & 7.60 & 1.70 \\
        \multicolumn{1}{l||}{\name} & \textbf{47.57} & \textbf{81.64} \\
        \midrule
        \multicolumn{1}{c||}{Model} & \multicolumn{1}{c}{Falcon} & \multicolumn{1}{r}{Size: 7B}\\
        \midrule
        \multicolumn{1}{l||}{FT*} & 5.60 & 17.20 \\
        \multicolumn{1}{l||}{ROME*} & 1.70 & 7.30 \\
        \multicolumn{1}{l||}{MEMIT*} & 2.30 & 1.60 \\
        \multicolumn{1}{l||}{\name} & \textbf{39.10} & \textbf{81.69} \\
        \midrule
    \end{tabular}
    \caption{Per-case accuracy of  compared against parameter-based knowledge editors. Approaches marked with (*) indicate results from a previous work.}
    \label{tab:vs_older}
\end{table}